\documentclass[11pt]{article}

\usepackage[final]{acl}

\usepackage{times}
\usepackage{latexsym}

\usepackage[T1]{fontenc}
\usepackage[utf8]{inputenc}

\usepackage{microtype}

\usepackage{inconsolata}

\usepackage{graphicx}
\usepackage{tcolorbox}
\usepackage{booktabs}
\usepackage{multirow}

\title{ShiZhi: A Chinese Lightweight Large Language Model for Court View Generation}

\author{
    \textbf{Zhitian Hou\textsuperscript{1}},
    \textbf{Kun Zeng\textsuperscript{1,*}}
\\
    \textsuperscript{1}School of Computer Science and Engineering, Sun Yat-sen University, China \\
    \textsuperscript{*}Corresponding Author
\\
    \small{
       \textbf{Correspondence:} \href{mailto:houzht@mail2.sysu.edu.cn}{houzht@mail2.sysu.edu.cn};
       \href{mailto:zengkun2@mail.sysu.edu.cn}{zengkun2@mail.sysu.edu.cn}
   }
}

\begin{document}
\maketitle
\begin{abstract}
Criminal Court View Generation (CVG) is a fundamental task in legal artificial intelligence, aiming to automatically generate the "Court View" section of a legal case document. Generating court views is challenging due to the diversity and complexity of case facts, and directly generating from raw facts may limit performance. In this paper, we present ShiZhi, the first large language model (LLM) specifically designed for court view generation. We construct a Chinese Court View Generation dataset, CCVG, of more than 110K cases, each containing fact descriptions paired with corresponding court views. Based on this dataset, ShiZhi achieving 70.00 ROUGE-1 and 67.85 BLEU-1 on court view generation, as well as 86.48\% accuracy with 92.75\% macro F1 on charge prediction. Experimental results demonstrate that even a small LLM can generate reasonable and legally coherent court views when trained on high-quality domain-specific data. Our model and dataset are available at \href{https://github.com/ZhitianHou/ShiZhi}{https://github.com/ZhitianHou/ShiZhi}.
\end{abstract}

\section{Introduction}
In recent years, Legal Artificial Intelligence (Legal AI) has gained significant attention due to its potential to assist in judicial decision-making, legal document analysis, and other tasks \citep{hou2025largelanguagemodelsmeet}. Among various tasks in Legal AI, Criminal Court View Generation (CVG) has emerged as an important problem\citep{ye-etal-2018-interpretable}. Specifically, in countries with a case law system, such as the United States, legal case documents generally consist of Procedure, Fact, Court View, Decision, and Tail sections. In contrast, the structure of Chinese legal case documents is implicitly conveyed through text formatting\citep{li2023sailer}. Moreover, criminal court views typically comprise rationales and charges, where the charges are derived and explained based on the rationales\citep{wu2020biased, ye-etal-2018-interpretable}. The goal of CVG is generating "Court View" section of legal case documents based on Fact.

\begin{figure}[t]
  \centering
  \includegraphics[width=\columnwidth]{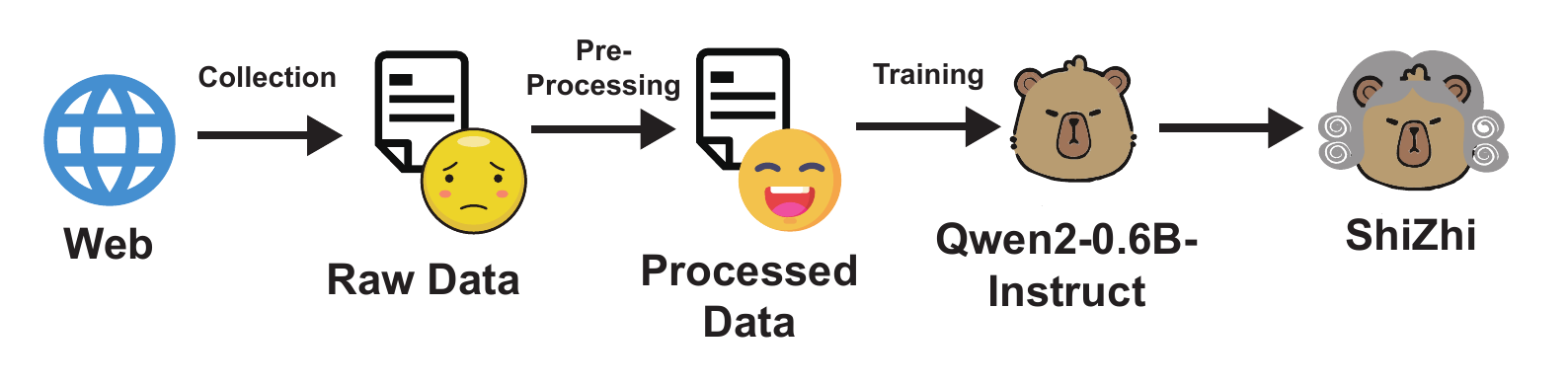}
  \caption{The pipeline of our data curation and model training.}
  \label{fig:pipeline}
\end{figure}

\begin{figure*}[t]
  \centering
  \includegraphics[width=0.8\linewidth]{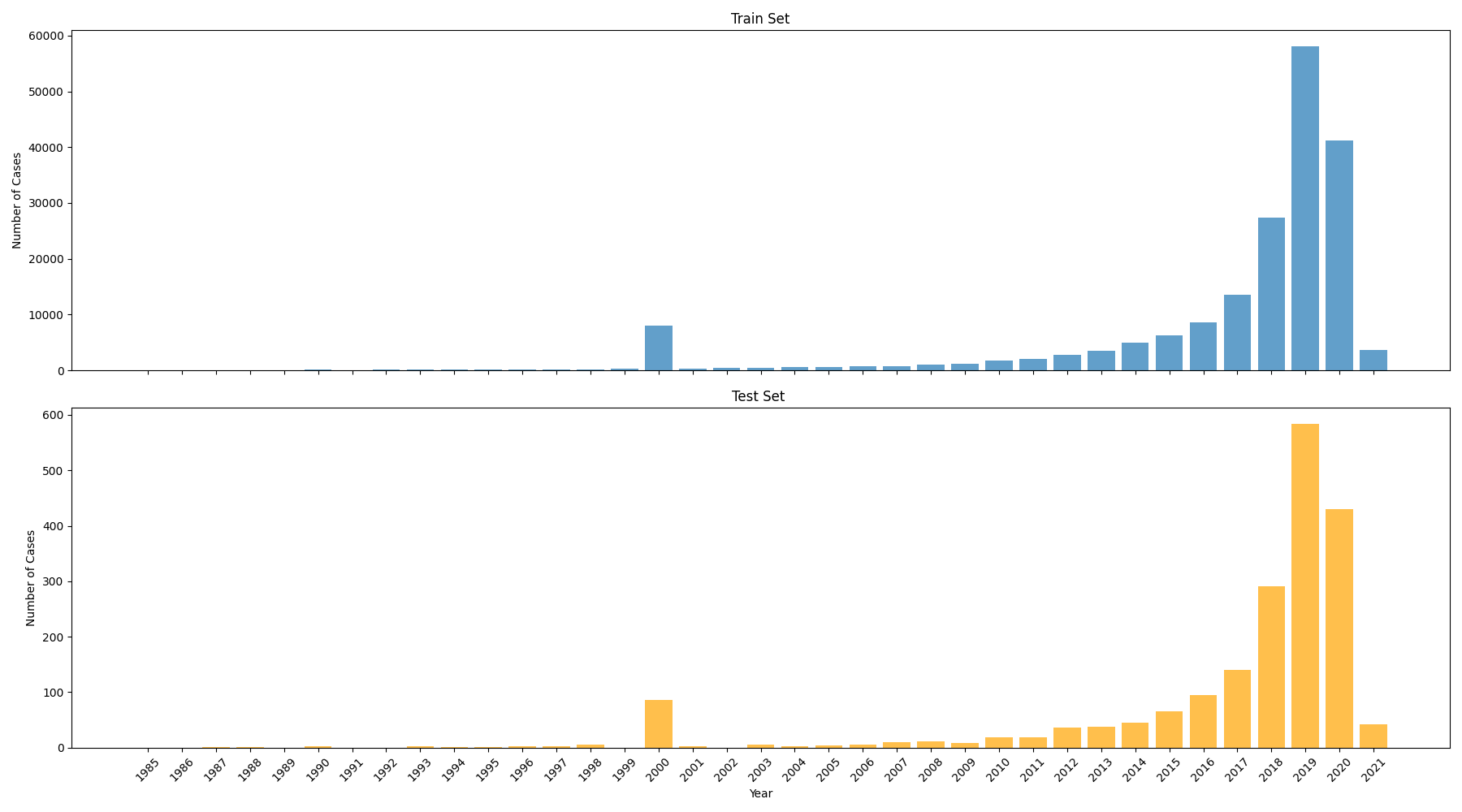}
  \caption{The distribution of case occurrence years in the train and test sets.}
  \label{fig:year statistic}
\end{figure*}

A number of recent works have explored neural network-based approaches for CVG. Yue et al. \citep{yue2021circumstances} designed a circumstances-enhanced framework to separately generate adjudicating and sentencing reasoning. Yue et al. \citep{yue2024event} proposed Event Grounded Generation (EGG), which introduces fine-grained event information from case facts into court view generation. EGG first extracts events from case facts using LLM-based methods and incorporates them into the generation by merging facts and events. Despite these efforts, CVG remains a text-to-text generation task, making it a natural fit for large language models (LLMs). However, to date, there is no dedicated LLM specifically trained for court view generation.

To fill this gap, we develop ShiZhi, a 0.5B-parameter LLM for CVG. The model size is comparable to Pretrained Language Models (PLMs) widely used in CVG research. Our contributions are threefold:
\begin{itemize}
    \item We curate a high-quality Chinese dataset, \textbf{CCVG}, specifically for court view generation.
    \item We fine-tune a 0.5B LLM, \textbf{ShiZhi}, on CCVG dataset for the CVG task.
    \item We demonstrate that even a lightweight LLM can generate reasonable and legally coherent court views, achieving strong performance on this task.
\end{itemize}

\section{Related Work}
\noindent \textbf{Structured Modeling for Court View Generation.}
Prior studies have proposed various techniques to enhance the generation of court views by incorporating structured legal information. For example, some works focus on extracting fine-grained legal features such as adjudicating and sentencing circumstances or legal concepts \citep{yue2021circumstances, xu2024divide} from case facts to improve the informativeness and faithfulness of generated rationales. Others introduce knowledge-aware mechanisms, such as injecting external law articles, charges, and claim information \citep{ye2018interpretable, li2021court} or incorporating legal knowledge bases through prompt tuning and guidance \citep{li2024enhancing}. To further address the interpretability and fairness of generated texts, several works employ causality-based reasoning, such as counterfactual generation \citep{wu2020biased, huang2023improving}, or introduce modular generation with QA-based slot filling \citep{huang2021generating}. These approaches demonstrate the effectiveness of integrating legal structure and knowledge into the generation process.

\noindent \textbf{Leveraging Large Language Models in Generation.} With the advancement of large-scale PLMs, recent research has begun to explore their potential in court view generation. Some works use general-purpose LLMs to extract intermediate legal structures \citep{yue2024event}, while others explore ways to stimulate internal legal knowledge or inject external guidance to improve performance in knowledge-intensive legal domains \citep{liu2024unleashing}. Although these methods demonstrate the feasibility of using LLMs in CVG, they mostly rely on general LLMs designed for open-domain tasks and lack models specifically trained for court view generation. 

Compared to existing methods, our model integrates domain-specific legal reasoning into an end-to-end generation pipeline, offering a compact and efficient solution for CVG.

\begin{figure*}[h]
  \centering
  \includegraphics[width=0.7\linewidth]{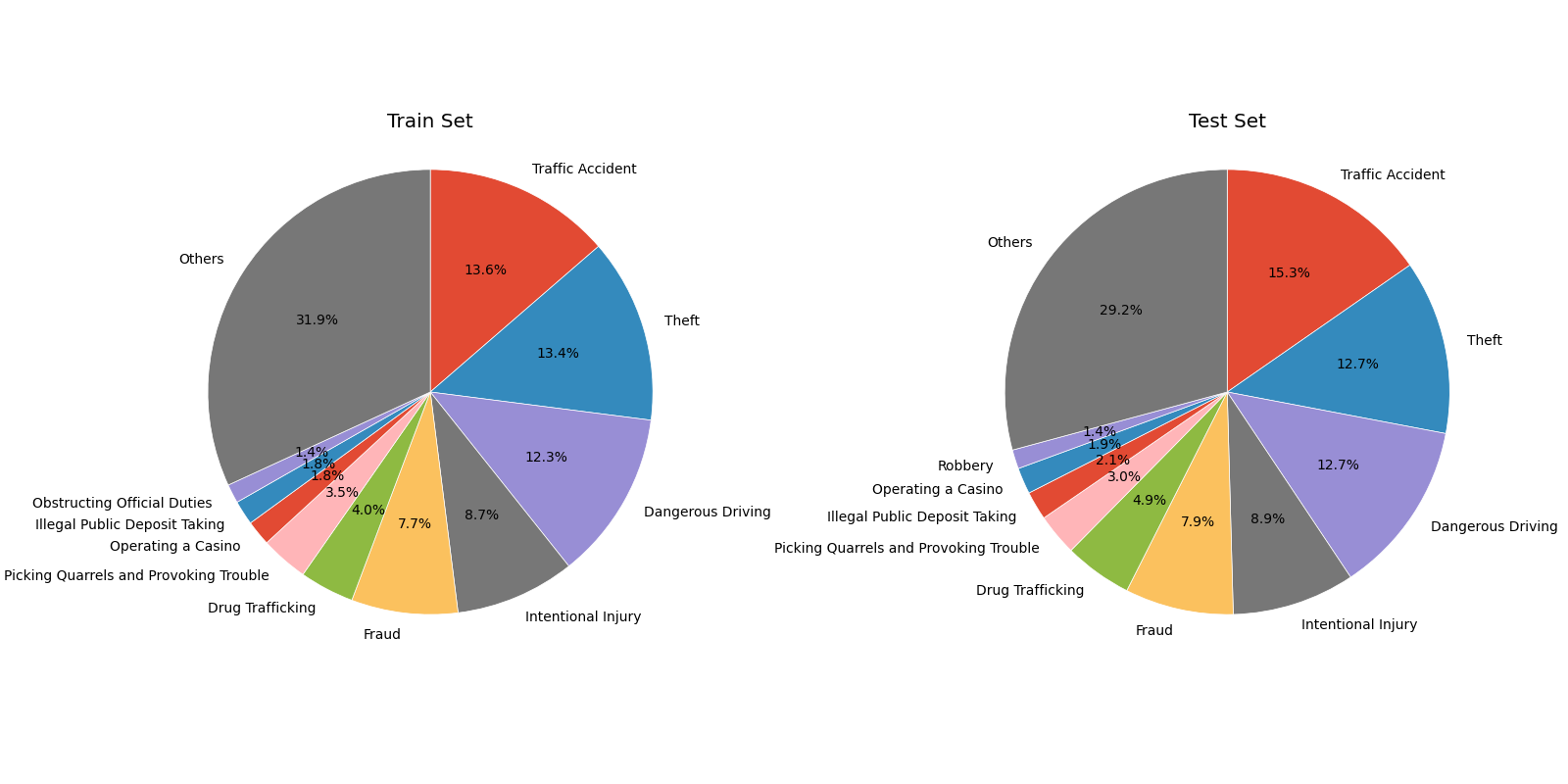}
  \caption{The distribution of charges in train and test sets.}
  \label{fig:charge statistic}
\end{figure*}

\begin{table*}[t]
  \centering
  \begin{tabular}{lccccc}
    \hline
    \textbf{Subset} & \textbf{Min} & \textbf{Max} & \textbf{Mean} & \textbf{Std} & \textbf{Median} \\
    \hline
    Train-Fact & 213 & 1013 & 417.9 & 175.2 & 367  \\
    Train-Court View & 200 & 1000 & 300.9 & 119.49 & 261  \\
    Test-Fact & 213 & 1013 & 408.1 & 161.6 & 365 \\
    Test-Court View & 200 & 979 & 299.5 & 116.3 & 260 \\
    \hline
  \end{tabular}
  \caption{The descriptive statistics of length of fact and court view in train and test sets.}
  \label{tab:descriptive statistics}
\end{table*}

\section{Data Curation}
Our data curation and model training pipeline is illustrated in Figure ~\ref{fig:pipeline}. We begin by collecting Chinese legal case documents from China Judgments Online\footnote{https://wenshu.court.gov.cn/} spanning from 1985 to 2021. These legal case documents contain rich fact descriptions and court views, which form the basis for our court view generation dataset, CCVG. To construct a high-quality dataset tailored for CVG, we apply a multi-step filtering and preprocessing process.

\subsection{Section Extraction}
Unlike legal documents in common law systems, Chinese legal case documents do not explicitly separate sections such as Facts and Court Views using structural markers. Instead, they are often implied through standard phrases. The Fact section typically begins with phrases such as \textit{"after identification"}. The Court View section often starts with \textit{"the court holds that"}. The Decision section generally follows \textit{"in accordance with the law"}. Using these lexical cues, we design a set of regular expressions to extract the Fact and Court View sections for each document. If either the fact or the court view section fails to be successfully extracted, the corresponding sample is discarded. We also split the dataset into train and test sets with a 9:1 ratio. The distribution of case occurrence years in the train and test sets is shown in the Figure \ref{fig:year statistic}.

\subsection{Charge Extraction}
In addition, we also extracted the charge type using regular expressions. However, since our primary objective is court view generation, a small portion (0.2\%) of the training samples which did not have the charge field successfully extracted are remained. For these samples, we set the "charge" field to \textit{null}. Nevertheless, we ensured that all fields in the test set are non-empty. The distribution of charges is shown in the Figure \ref{fig:charge statistic}.

\subsection{Post Filtering}
\begin{figure}[h]
  \centering
  \includegraphics[width=\columnwidth]{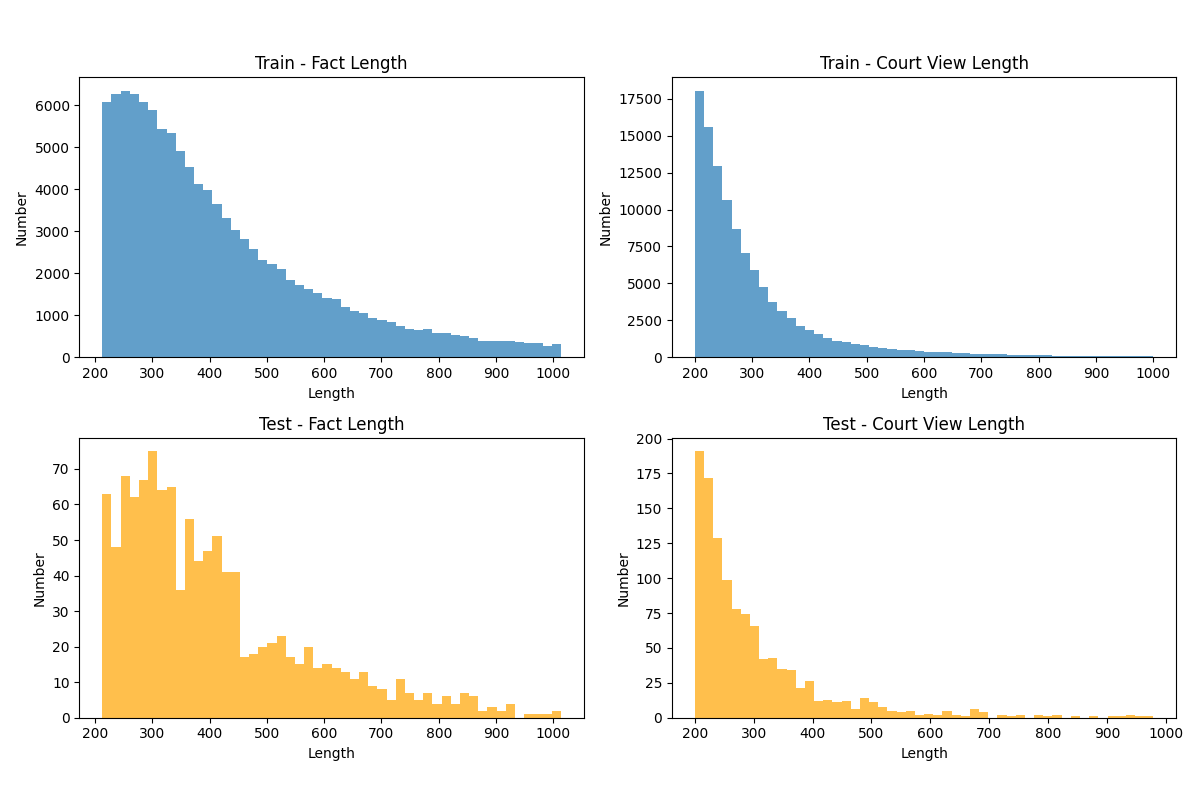}
  \caption{The length statistic of fact and court view in train and test sets.}
  \label{fig:length statistic}
\end{figure}

To ensure data quality and compatibility with our model architecture, we apply length-based filtering. We discard any sample where either the fact or court view text is shorter than 50 characters or longer than 1024 characters after extraction, which helps reduce noise and computational inefficiency. The length statistic of fact and court view in train and test sets are shown in Figure \ref{fig:length statistic} and their descriptive statistics presented in the Table \ref{tab:descriptive statistics}.

\begin{table*}[t]
  \centering
  \begin{tabular}{c|ccc|ccc|cc}
    \toprule
    \multirow{4}{*}{Models} & \multicolumn{6}{c|}{Court View Generation} & \multicolumn{2}{c}{Charge} \\
    \cmidrule(lr){2-7}
    & \multicolumn{3}{c|}{ROUGE} & \multicolumn{3}{c|}{BLEU} & \multicolumn{2}{c}{Prediction} \\
    \cmidrule(lr){2-4} \cmidrule(lr){5-7} \cmidrule(lr){8-9}
    & R-1 & R-2 & R-L & B-1 & B-2 & B-N & Acc & MF1 \\
    \midrule
    Qwen2-0.5B-Instruct & 24.18 & 6.53 & 16.40 & 29.16 & 17.11 & 9.24 & 16.29 & 28.02\\
    \textbf{ShiZhi} & \textbf{70.00} & \textbf{51.07} & \textbf{59.61} & \textbf{67.85} & \textbf{62.92} & \textbf{54.76} & \textbf{86.48} & \textbf{92.75} \\
    \bottomrule
  \end{tabular}
  \caption{Results of court view generation and the charge prediction.}
  \label{tab:court_view_charge_results}
\end{table*}

The final dataset consists of high-quality, structured document pairs (fact, ~ court view), which we use to fine-tune our language model for the CVG task.

\section{Model Training}
We trained our model using the Swift framework \citep{zhao2025swift}, an efficient and flexible toolkit designed for instruction-tuning and fine-tuning LLMs. Swift supports various backends and model families, and offers lightweight abstractions for managing datasets, prompt templates, and training loops, which makes it particularly suitable for rapid prototyping and deployment in legal NLP applications.

ShiZhi is training based on Qwen2-0.5B-Instruct, a lightweight instruction-tuned Chinese language model with strong performance. We fine-tuned it on our curated dataset, CCVG, using a prompt format tailored for the CVG task. Specifically, we constructed a new system prompt that guides the model to behave as a legal assistant. The fact description is treated as the query and the court view is the expected response as prompt template.

\begin{tcolorbox}[
    colback=blue!6!white,
    colframe=black,
    colbacktitle=black,
    coltitle=white,
    fonttitle=\bfseries\sffamily,
    title=Prompt Template,
    sharp corners,
    boxrule=1pt,
]
\{"system": “Assume the role of a judge. Based on the fact section of a legal case document, generate the corresponding court view section.”, "query": Fact description:\textbackslash n \{fact\}\textbackslash n Court View:", "response": \{court view\}, "charge": \{charge\}\}
\end{tcolorbox}

The model was trained for 3 epochs with a batch size of 4 and a learning rate of 1e-3. The maximum input length was set to 1024 tokens. Training was conducted on a single NVIDIA RTX 4090 GPU.

\section{Experiment Results}
\subsection{Evaluation Metrics}
We evaluated our model using a combination of text generation metrics and charge prediction metrics to comprehensively assess the quality of the generated court views. For text generation, we compute ROUGE-1, ROUGE-2, and ROUGE-L to measure unigram, bigram, and longest common subsequence overlaps between the generated court view and the label, and BLEU-1, BLEU-2, and BLEU-n to assess n-gram precision. To evaluate charge prediction correctness, we measure F1 score and accuracy (Acc). Specifically, for each test case, we check whether the correct charge label appears in the generated court view; if the charge is present, it is counted as correct, otherwise it is considered incorrect. F1 score and accuracy are then computed over the entire test set. This combination of metrics allows us to capture both the linguistic quality and legal fidelity of the generated court views.

\subsection{Main Results}
Table~\ref{tab:court_view_charge_results} reports the performance of different models on court view generation and charge prediction. Among the compared models, ShiZhi consistently outperforms the baseline Qwen2-0.5B-Instruct across all metrics. For court view generation, ShiZhi achieves a ROUGE-1 of 70.00, ROUGE-2 of 51.07, and ROUGE-L of 59.61, substantially higher than the baseline scores of 24.18, 6.53, and 16.40, respectively. Similarly, BLEU scores are markedly improved, with BLEU-1 at 67.85 compared to 29.16 for the baseline. The large gap between ShiZhi and the baseline demonstrates the effectiveness of domain-specific training for capturing legal reasoning and structuring coherent court views.

For charge prediction, ShiZhi achieves 86.48\% accuracy and 92.75\% macro F1, substantially surpassing the baseline performance of 16.29\% and 28.02\%, respectively. These results indicate that domain-specific fine-tuning allows ShiZhi to effectively learn the mapping from fact descriptions to charges, reflecting strong legal reasoning capabilities.

Overall, the results demonstrate that curating a specialized dataset and fine-tuning a lightweight LLM enables effective generation of court views and accurate charge prediction, while highlighting the challenges of evaluation in highly variable legal text.

\section{Conclusion}
In this paper, we presented \textbf{ShiZhi}, the first large language model specifically designed for Criminal Court View Generation (CVG). We curated \textbf{CCVG}, a dataset of over 110K Chinese criminal cases with fact descriptions paired with court views, and fine-tuned a 0.5B lightweight LLM on this dataset. ShiZhi achieves 70.00 ROUGE-1 and 67.85 BLEU-1 on court view generation, as well as 86.48\% accuracy with 92.75\% macro F1 on charge prediction, demonstrating that even a small LLM can generate legally coherent and factually grounded court views when trained on high-quality domain-specific data. Our model and dataset provide a foundation for future research in automated legal document generation and legal reasoning with LLMs.

\section*{Limitations}
Despite the effectiveness of ShiZhi, there are several limitations in the current work. First, our model and dataset are exclusively in Chinese, limiting its applicability to other languages and legal systems. Second, the curated dataset, CCVG, only includes cases up to 2021, which may not reflect the most recent legal developments and case patterns. Third, we only explored a single parameter scale (0.5B) for the LLM, without investigating the impact of different model sizes on performance. In future work, we plan to leverage larger and more advanced LLMs as well as updated datasets, in order to study their capabilities on CVG and charge prediction tasks, and to systematically analyze how LLMs understand and reason over legal content.


\bibliography{custom}

\end{document}